\documentclass[conference]{IEEEtran}
    \IEEEoverridecommandlockouts
    
    \usepackage{cite}
    \usepackage{amsmath,amssymb,amsfonts}
    \usepackage{algorithmic}
    \usepackage{graphicx}
    \usepackage{textcomp}
    \usepackage{xcolor}
    \usepackage{hyperref}
    \def\BibTeX{{\rm B\kern-.05em{\sc i\kern-.025em b}\kern-.08em
        T\kern-.1667em\lower.7ex\hbox{E}\kern-.125emX}}
\usepackage{xspace} 

\author{
\IEEEauthorblockN{Jonathan Yang}
\IEEEauthorblockA{\textit{Department of Computer Science} \\
\textit{Stanford University}\\
Stanford, CA, USA \\
}
\and
\IEEEauthorblockN{Chelsea Finn}
\IEEEauthorblockA{\textit{Department of Computer Science} \\
\textit{Stanford University}\\
Stanford, CA, USA \\
}
\and
\IEEEauthorblockN{Dorsa Sadigh}
\IEEEauthorblockA{\textit{Department of Computer Science} \\
\textit{Stanford University}\\
Stanford, CA, USA \\
}
}

\begin{document}
    
    \title{Cross-Controller Transfer via In-Context Behavior Adaptation}
    
    \maketitle

\begin{figure*}[t]
    \centering
    \includegraphics[width=\textwidth]{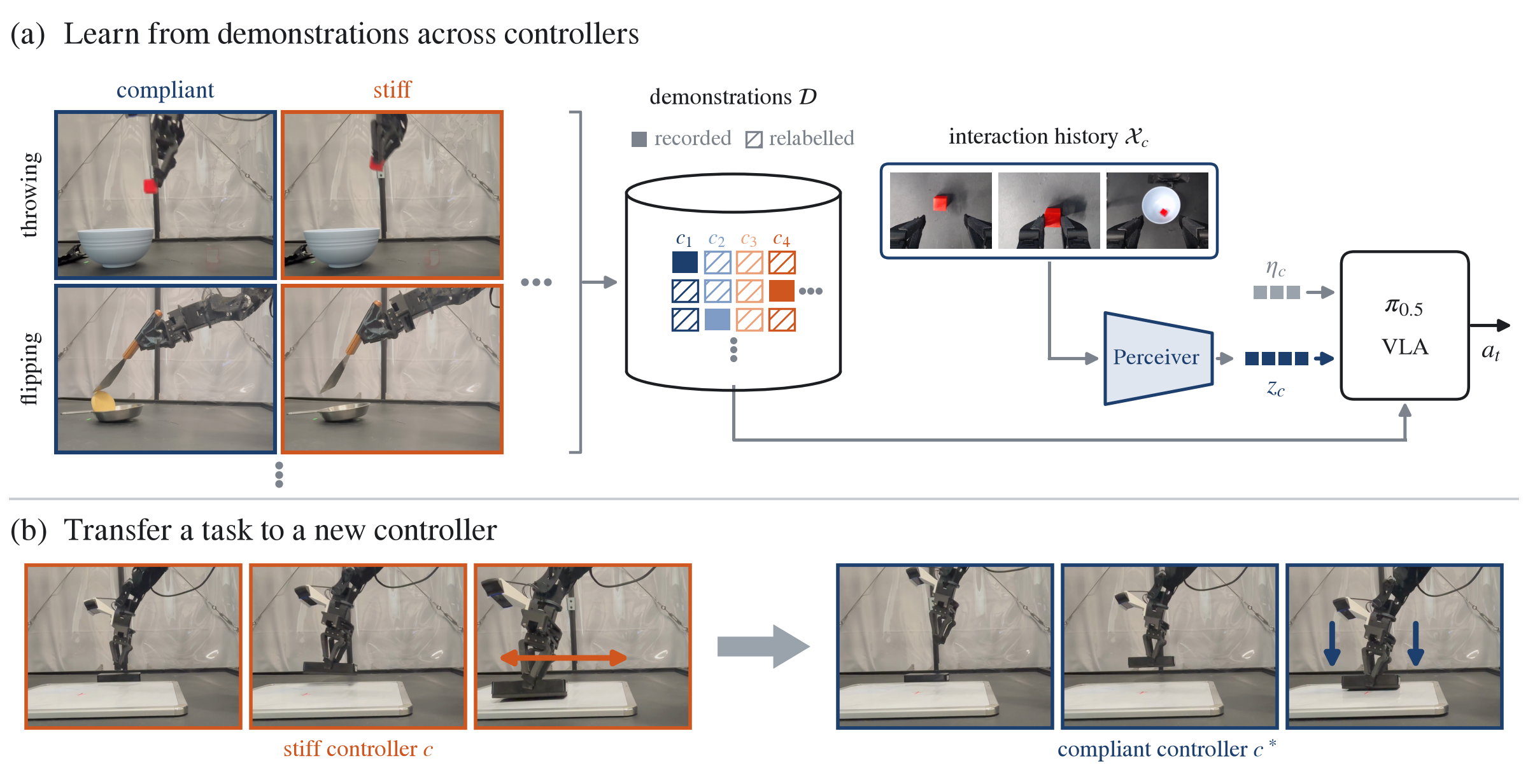}
    \caption{\textbf{Cross-Controller Transfer.} We train a generalist policy that adapts its manipulation strategy across a dense spectrum of unseen controller types and continuous gains by conditioning on interaction history.}
    \label{fig:teaser}
\end{figure*}
    
    \begin{abstract}
Generalist robot policies increasingly learn from data collected across diverse tasks and embodiments, yet typically assume a single controller for each robot. This assumption limits transfer because different controllers can assign different physical meanings to the same action and may require different strategies to complete a task. We study how to transfer robot data across both embodiments and controllers without allowing conflicting action interpretations to interfere with learning. Our setting spans six controller types as well as variations in controller gains and compliance. We propose a framework that encodes behavioral interaction under a controller through examples of commanded actions and the resulting robot motions. A transformer context encoder summarizes these interactions, allowing a single policy to adapt in context. Across real-world experiments, our method improves average success by 33 percentage points over naive data pooling, achieves 50\% success on unseen controller modes and 52\% on unseen controller gains, and transfers to tasks with limited controller-specific data. On contact-rich tasks such as USB insertion and surface wiping, behavioral interaction context enables the policy to adapt not only its actions but also its strategy. Our results provide a path toward generalist robot policies that can leverage data collected across diverse action interpretations. For more information, visit \href{https://controller-transfer.github.io}{https://controller-transfer.github.io}.
    \end{abstract}
    
    
\section{Introduction}
Deploying a policy on a new robot raises a deceptively simple question: \emph{how will its actions actually move the robot?} Scaling robot datasets has produced increasingly general policies that operate across diverse scenes, tasks, and embodiments. Yet most policies assume that large portions of their data share the same controller per robot and that the controller used at deployment is known a priori. This assumption is limiting: in reality, data contains significant variability across control spaces (e.g., joint vs. Cartesian), control targets (e.g., position vs. velocity), and control parameters (e.g., varying stiffness or gains). For example, data collected through VR teleoperation might encode compliant Cartesian commands, while data collected through puppeteering might encode stiff joint targets, causing the exact same action vector $\mathbf{a}$ to have substantially different physical interpretations. Moreover, even on the same embodiment, different tasks may benefit from different controllers. In this paper, we study the question: \emph{how can we learn from data collected across this wide range of controller configurations?} 
    
Two lines of prior work inform controller transfer, but address different parts of the problem. Cross-embodiment research has shown that data can transfer across robots with different morphologies, sensors, and action spaces, but typically assumes a single, fixed controller for each robot setup. For example, in large-scale multi-robot policies like RT-X\cite{oxe2024rtx}, a specific robot and its associated controller are uniquely tied to a distinct physical lab or background. Consequently, the policy can easily infer the controller setup purely from the visual scene, sidestepping the challenge of conflicting action interpretations. Meanwhile, research on compliant control shows that controller choice directly impacts task performance\cite{aljalbout2024role}. For example, data collected with a controller that feels natural to a human teleoperator may not produce the data most beneficial for policy learning \cite{bronars2026tune}. However, this line of work typically studies how to select or tune a controller for a particular robot and task, rather than how to learn from data collected under many controllers. Our work bridges these two directions: we study how a generalist policy can pool demonstration data with conflicting action interpretations and adapt its behavior when the active controller cannot be inferred from the visual scene.

At first glance, one might treat controller transfer as a dynamics problem: identify the active controller and analytically remap actions to match nominal forces or kinematics. However, dynamics remapping only preserves how an existing motion is realized under new dynamics, not how an expert would choose to act. Consider USB insertion: under a compliant controller, a human demonstrator deliberately presses against the port and searches laterally for alignment. Under a stiff controller, that same contact-seeking trajectory causes the connector to jam, requiring a cautious, alignment-first approach that avoids contact altogether. While system identification or torque matching can predict how a controller will respond to an action, it cannot invent the strategy an expert would adopt. Cross-controller transfer is therefore not a matter of remapping dynamics, but of inferring the appropriate behavioral strategy and determining the actions that will realize that strategy.

Learning controller-dependent strategies from controller parameters and the current observation alone would require dense demonstration coverage across task--controller combinations, which is costly to collect on physical robots. Instead, we condition the policy on interaction trajectories collected under the deployment controller, even when those trajectories come from other tasks. A transformer context encoder inspired by Perceiver~\cite{jaegle2021perceiverio} compresses these interactions into a fixed set of tokens that condition a single vision-language-action (VLA) policy. By learning across source tasks, the policy can reuse this behavioral interaction context to select an appropriate strategy for a target task with little or no target-task data under that controller. Finally, since physical demonstrations inevitably have limited controller coverage, we also introduce a synthetic data augmentation scheme that analytically remaps actions across controllers; the behavioral context disambiguates the resulting multimodal action modes, enabling safe scaling across diverse controllers.

We make three contributions. First, we formulate cross-controller transfer as behavioral adaptation under conflicting action interpretations, distinguishing the problem from system identification and action remapping. Second, we develop a context-conditioned VLA policy that combines synthetic controller relabeling with a Perceiver encoder, which compresses commanded--achieved interaction histories into fixed-length tokens for adaptation without test-time training. Third, we evaluate the approach across five real-world manipulation tasks, six controller types, and continuous gain variations. When tasks are demonstrated under only a single controller type, our method improves zero-shot success by $23$ percentage points over task-neutral dynamics context, improves success by $35$ percentage points over synthetic relabeling alone on unseen controller modes, and transfers behavioral context across contact-rich tasks while retaining $78\%$ of in-domain performance.

\section{Related Work}

\textbf{Cross-Embodiment Learning.}
Recent generalist policies have demonstrated positive tranfer scaling data by by aggregating demonstrations across diverse robots, scenes, and tasks~\cite{brohan2022rt, brohan2023rt2, oxe2024rtx, walke2023bridgedata, khazatsky2024droid, kim2024openvla, black2024pi0, intelligence2025pi05}. To accommodate varying kinematic limits and command formats, architectures such as CrossFormer~\cite{doshi2024crossformer} and Octo~\cite{octo2024} use embodiment-specific action heads and encoders, while UniVLA~\cite{bu2025univla} learns latent actions decoded into platform-specific controls. While these approaches bridge structural differences across action \emph{spaces}, they treat the underlying controller for each embodiment as fixed, implicitly assuming the control stack can be inferred from the visual scene. At scale, this assumption breaks down: when the same robot uses different controllers across tasks or teleoperation setups, the visual scene no longer identifies how commands are realized, leading to conflicting action interpretations.

\textbf{Controller Adaptation and Policy Learning.}
A complementary line of research studies how low-level controllers and action-space representations govern policy learning and physical interaction~\cite{hogan1985impedance, khatib1987operational, mason1981compliance, varin2019actionspaces, martin2019vices, aljalbout2024role}. These works demonstrate that controller choice and compliance strongly shape learnability, with compliant impedance control outperforming stiff position control in contact-rich tasks~\cite{bronars2026tune}. Recent methods further expand this capability by making compliance practical on commodity arms without dedicated force-torque sensing~\cite{shi2026minimalist, hou2025adaptive, xu2025compliant}. However, this body of work almost universally studies controllers in isolation: selecting, tuning, or learning a specific controller for a single robot and task. A related direction conditions policies on execution factors such as speed~\cite{jing2026tempovla}, showing that a single policy can flexibly modulate how it acts. However, these works do not address how a policy can learn from datasets pooled across heterogeneous controllers.

\textbf{Online Policy Adaptation.}
Our work is closely related to a learning-based line that adapts policies to physical variation without re-tuning. System identification and dynamics randomization~\cite{peng2018simtoreal} help policies accommodate changes such as terrain friction or payload; however, they target plant dynamics rather than manipulation strategy. In parallel, in-context and memory-augmented policies~\cite{duan2017oneshot, finn2017oneshot, torne2026mem} adapt behavior directly from past trajectory context without retraining. Our method bridges both: rather than identify physical parameters, we cast cross-controller transfer as in-context policy conditioning, compressing a short history of commanded-versus-achieved transitions so that a generalist VLA infers how the deployment controller behaves and adapts its manipulation strategy accordingly.

\begin{figure}[t]
    \centering
    \includegraphics[width=\columnwidth]{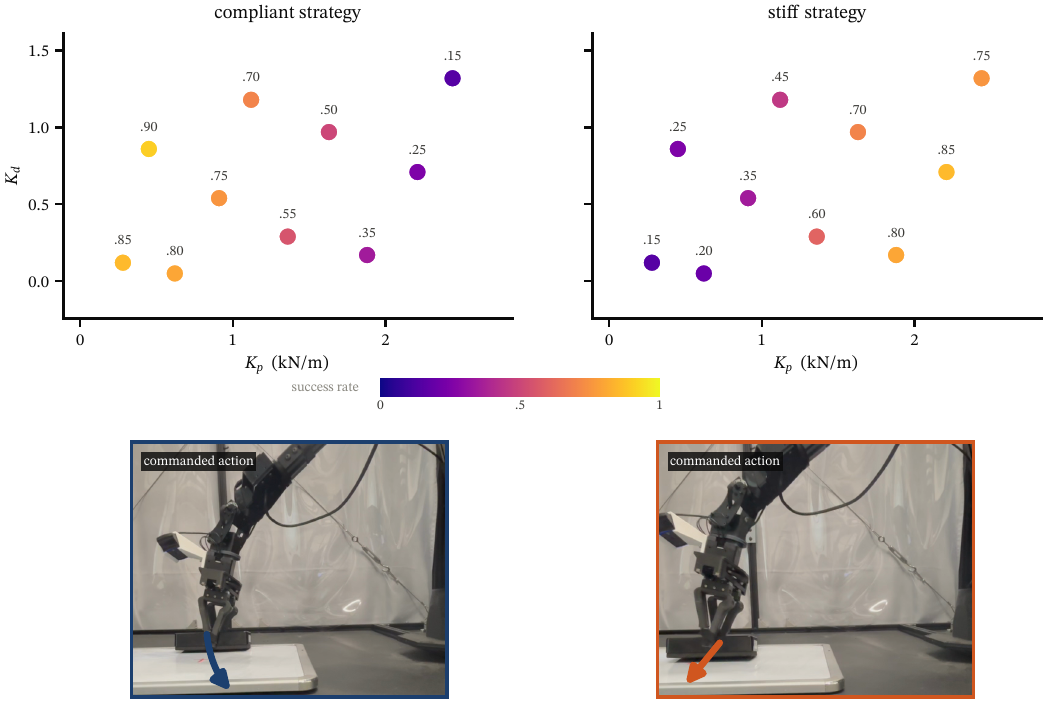}
    \caption{\textbf{The optimal strategy differs per controller.} Two learned wiping strategies evaluated across a shared $(K_p,K_d)$ plane. Low stiffness favors compliant pressing ($80\%$--$90\%$), whereas high stiffness favors stiff tracing ($75\%$--$85\%$).}
    \label{fig:strategy_gain_compatibility}
\end{figure}

\section{Understanding Controller Variation}

\label{sec:understanding_controllers}

This section characterizes controller variation and system identification. We then consider whether known controller dynamics can be used to transfer demonstrations, and show why such transfer does not recover expert behavior.

\subsection{What Is a Robot Controller?}
\label{subsec:what_is_a_controller}

Robot policies rarely command raw motor torques directly. Because manipulator dynamics requires feedback loops ($500$\,Hz--$1$\,kHz) far faster than learned policy inference ($10$--$50$\,Hz), policies instead emit reference setpoints $a_t \in \mathcal{A}_c$ (e.g., target positions or velocities). A low-level controller converts these commands into actuator torques:
\begin{equation}
    \tau_t = g_c(a_t, q_t, \dot{q}_t; \theta_c),
    \label{eq:controller_mapping}
\end{equation}
where $q_t \in \mathcal{Q}$ and $\dot{q}_t \in \dot{\mathcal{Q}}$ denote joint positions and velocities, $g_c$ is the control law, and $\theta_c$ denotes its governing parameters (e.g., feedback gains and stiffness). We refer to the tuple $c = (\mathcal{A}_c, g_c, \theta_c)$ as the \emph{controller configuration}, which induces closed-loop transition dynamics $p_c(q_{t+1} \mid q_t, \dot{q}_t, a_t)$.

\subsection{Controller Identification and Dynamics Transfer}
\label{subsec:controller_identification}

When deploying a policy across diverse robotic platforms or setups, the deployment controller configuration is typically unknown or differs from the demonstration environment. A natural first inclination is to treat cross-controller adaptation as a classical system identification and dynamics problem: calibrate the active controller from observed motion, and analytically remap demonstrated commands so that the target controller produces equivalent actuator torques.

Specifically, when the controller configuration is unknown, it can be inferred from a calibration history $\mathcal{H}_c = \{(q_t,\dot{q}_t,a_t,q_{t+1})\}_{t=1}^T$ of commanded references and achieved motion. System identification (SysID) fits a model that predicts the observed transitions:
\begin{equation}
    \hat{p}_c = \arg\max_{p} \sum_{t=1}^{T}
    \log p(q_{t+1} \mid q_t,\dot{q}_t,a_t).
\end{equation}
When the controller structure $g_c$ is known, this reduces to estimating its governing parameters $\hat{\theta}_c$; otherwise, $\hat{p}_c$ can be represented as a black-box forward model. \\
Once the source and target control laws (or their estimated parameters $\hat{\theta}_c, \hat{\theta}_{c'}$) are known, one can algebraically remap demonstrated commands so that both controllers produce the same torques. For instance, consider a standard proportional-derivative (PD) joint impedance control law (shared by all of our considered controller types):
\begin{equation}
    \tau_t = K (q_{d, t} - q_t) + D (\dot{q}_{d, t} - \dot{q}_t) + \tau_{\mathrm{ff}}(q_t, \dot{q}_t),
    \label{eq:pd_law}
\end{equation}
where $a_t = (q_{d, t}, \dot{q}_{d, t})$ is the reference command, $K$ and $D$ are proportional and derivative gain matrices, and $\tau_{\mathrm{ff}}$ is gravity and Coriolis compensation. To transfer a recorded demonstration $(q_t, \dot{q}_t, a_t)$ to a target controller $c'$ with parameters $(K', D')$, one can solve for an equivalent command $a'_t = (q'_{d, t}, \dot{q}'_{d, t})$ that preserves instantaneous joint torque $\tau'_t = \tau_t$. Holding the reference velocity fixed ($\dot{q}'_{d, t} = \dot{q}_{d, t}$) yields:
\begin{equation}
    q'_{d, t} = q_t + (K')^{-1} \Big[ K (q_{d, t} - q_t) + (D - D')(\dot{q}_{d, t} - \dot{q}_t) \Big].
    \label{eq:relabeling}
\end{equation}

\begin{figure*}[t]
    \centering
    \includegraphics[width=\textwidth]{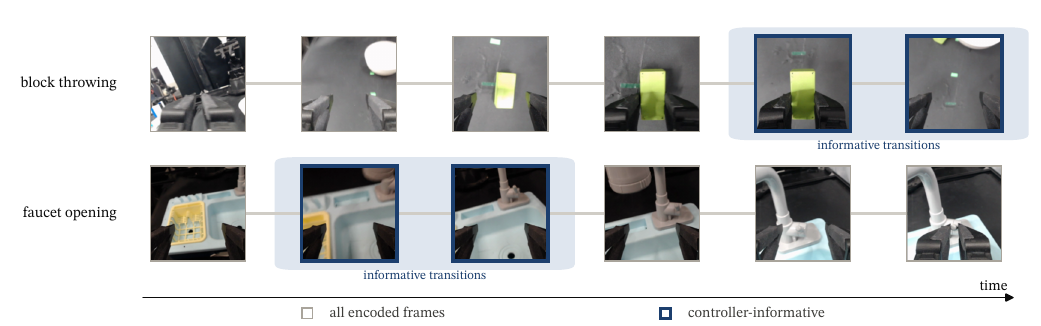}
    \caption{\textbf{Controller-informative events are not temporally aligned.} We encode every commanded--achieved transition in each trajectory; blue windows indicate where the strongest controller-relevant evidence occurs. Contact, braking, saturation, and recovery appear at different times across tasks, motivating complete trajectory context rather than a fixed temporal crop.}
    \label{fig:context_timeline}
\end{figure*}

\subsection{Dynamics Transfer Does Not Recover Expert Behavior}
\label{subsec:dynamics_gap}

While the algebraic remapping described above preserves instantaneous actuator torques, it only determines how an \emph{existing} motion is realized under new dynamics. It does not recover not how an expert demonstrator would choose to act under those dynamics. At the core of this breakdown is a fundamental distinction between two mathematical objects:
\begin{align}
    \text{closed-loop dynamics:} \quad & p_c(o_{t+1} \mid o_t, a_t), \\
    \text{expert behavior:} \quad & p(a_t \mid o_t, \ell, c),
\end{align}
where $\ell$ denotes the manipulation task. System identification and dynamics remapping estimate or manipulate the forward transition dynamics $p_c(o_{t+1} \mid o_t, a_t)$. However, knowing how a controller transforms actions does not dictate which actions a expert demonstrator selects: the physical dynamics $p_c$ do not uniquely specify expert behavior $p(a_t \mid o_t, \ell, c)$ without a model of behavioral adaptation. An action remapper can reproduce recorded forces along an existing path, but it cannot invent the trajectory that an expert would have executed under the new controller.

\section{Cross-Controller Policy Adaptation}
\label{sec:methods}

Motivated by the gap between controller dynamics and expert behavior, we cast cross-controller adaptation as a contextual learning problem. We train a policy that combines nominal controller metadata with a compact representation of behavioral interaction history, enabling it to adapt its behavior to be compatible with its controller.

To formalize this setting, we consider imitation learning from demonstrations pooled across diverse manipulation tasks $\ell \in \mathcal{L}$ and controllers $c \in \mathcal{C}$. At each timestep $t$, the policy observes $o_t = (I_t, q_t)$ comprising camera images $I_t$ and proprioception $q_t$, and predicts a chunk of actions $a_{t:t+H} \in \mathcal{A}_c^H$ over horizon $H$. Demonstrations are collected across a set of task--controller pairs $\mathcal{S}_{\mathrm{train}} \subset \mathcal{L} \times \mathcal{C}$, yielding a pooled dataset:
\begin{equation}
    \mathcal{D} = \bigcup_{(\ell, c) \in \mathcal{S}_{\mathrm{train}}} \mathcal{D}_{\ell, c},
\end{equation}
where each trajectory $\tau \in \mathcal{D}_{\ell, c}$ consists of sequence $\big((o_1, a_1), \dots, (o_T, a_T)\big)$. Because $\mathcal{S}_{\mathrm{train}}$ is sparse, most tasks are demonstrated under only a single controller configuration.

Under naive behavioral cloning on $\mathcal{D}$, the policy attempts to fit an unconditioned mixture over incompatible experts, $\sum_c p(c \mid o_t, \ell) \pi^\star_{\ell, c}$. Because standard robot observations $o_t$ do not identify the active controller, the policy averages across conflicting demonstrator strategies. Our objective is to learn a single generalist policy that can adapt to any target controller $c^\star \in \mathcal{C}$ at deployment—including novel controllers unseen during training—and recover the performance of the per-controller specialist $\pi^\star_{\ell, c^\star}$:
\begin{equation}
    \pi_\theta(a_{t:t+H} \mid o_t, \ell, \eta_{c^\star}, \mathcal{X}_{\ell_s,c^\star}) \;\approx\; \pi^\star_{\ell, c^\star}(a_{t:t+H} \mid o_t),
\end{equation}
where $\eta_{c^\star}$ is nominal metadata and $\mathcal{X}_{\ell_s,c^\star}$ is a small set of interaction trajectories collected on a source task $\ell_s$ under the target controller. The source task may equal the target task during training or differ from it during cross-task evaluation. Below, we introduce our context representation, encoder architecture, and training scheme.

Figure~\ref{fig:strategy_gain_compatibility} illustrates why the learned policy must account for controller-dependent behavior. Neither wiping strategy dominates globally: compliant pressing excels at low stiffness while stiff cautious-tracing achieves only $15\%$--$25\%$. At high stiffness, a less exploratory strategy instead dominates ($75\%$--$85\%$) while compliant pressing overloads under excessive contact forces. The learning problem is therefore not simply to identify the controller gains, but to select behavior compatible with them.

\subsection{Behavioral Interaction Context}
\label{subsec:behavioral_context}

\begin{figure*}[t]
    \centering
    \includegraphics[width=\textwidth]{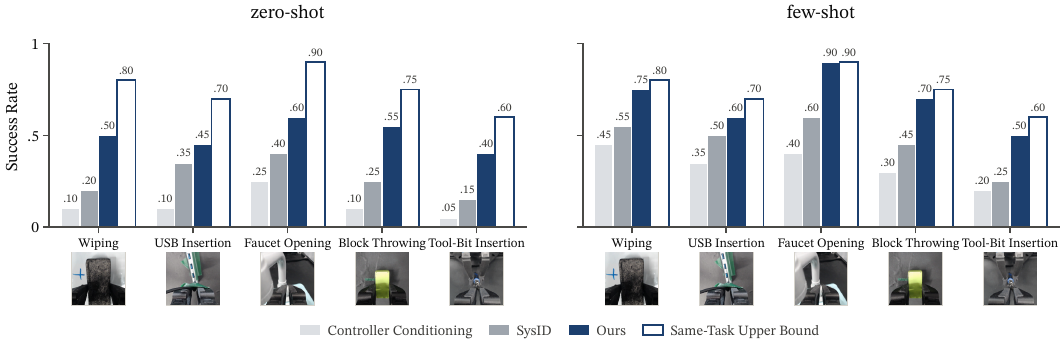}
    \caption{\textbf{Cross-task behavioral context provides information beyond controller dynamics.} Evaluation on held-out task--controller cells across five manipulation tasks ($N=20$ trials per cell). \textbf{Zero-shot} evaluates transfer where the target task has no demonstrations under deployment controller relying solely on target-task data collected under another controller. \textbf{Few-shot} provides a budget of $5$ demonstrations from the target (task, controller) pair.}
    \label{fig:cross_task_sysid}
\end{figure*}

An alternative to dynamics relabeling is conditioning the policy directly on estimated controller parameters $\hat{\theta}_c = (K, D)$. However, this requires the policy to infer how an expert would behave from abstract numerical gains alone. Learning this mapping would require dense demonstration coverage across task–gain combinations, which is prohibitively expensive on physical hardware. Instead, we condition the policy on two complementary sources of context: nominal controller metadata $\eta_c$ (a natural-language string specifying the controller type and gains) and an empirical set of interaction trajectories $\mathcal{X}_{\ell_s,c}$ recorded on task $\ell_s$ and controller $c$:
\begin{equation}
    \pi_\theta(a_{t:t+H} \mid o_t, \ell, \eta_c, \mathcal{X}_{\ell_s,c}).
    \label{eq:policy_context_conditioned}
\end{equation}
Consequently, $\mathcal{X}_{\ell_s,c}$ provides behavioral evidence generated under $c$; the target-task-conditioned policy determines which parts of that evidence are useful for its current decision. We represent behavioral interaction context as a set of $N$ trajectories:
\begin{equation}
    \mathcal{X}_{\ell_s,c} = \big\{ \tau_{\ell_s,c}^{(j)} \big\}_{j=1}^N, \quad \tau_{\ell_s,c}^{(j)} = \Big( \big(I_{j, i}, q_{j, i}, a_{j, i}^{\mathrm{cmd}}, q_{j, i+1}^{\mathrm{ach}}\big) \Big)_{i=1}^{M_j},
    \label{eq:transition_tuple}
\end{equation}
pairing camera images $I$ and commanded action references $a^{\mathrm{cmd}}$ with initial configurations $q$ and achieved robot configurations $q^{\mathrm{ach}}$.

\subsection{Context Encoder and Interaction Bottleneck}
\label{subsec:perceiver_resampler}

Encoding $\mathcal{X}_{\ell_s,c}$ presents two challenges. First, the learned representation must be reusable across tasks while remaining indicative of controller-mediated behavior. Second, controller-informative events occur sparsely and at different points across trajectories (Fig.~\ref{fig:context_timeline}), requiring broad temporal coverage without increasing the cost of inference. To fulfill these requirements, we encode interaction trajectories using a transformer with a fixed-size interaction bottleneck, similar to Perceiver architectures~\cite{jaegle2021perceiverio}. Rather than passing raw observation tokens directly to the policy, a set of learned query vectors attends over the interaction history to distill it into a fixed number of summary tokens. We do not assume that these tokens identify the controller or encode an explicit strategy; they provide evidence from controller-mediated behavior that the target-task-conditioned policy can use for action prediction.

\noindent\textbf{Encoder Architecture.} From each set of calibration interaction trajectories, we encode both visual scene context and physical transitions over a total context horizon of $T$ timesteps ($T=128$ by default; ablated over $T \in \{16, 32, 64, 128\}$ in Section~\ref{sec:experiments}). The sampled RGB camera frames are encoded with a frozen DINOv2 ViT-B/14 backbone~\cite{oquab2023dinov2}, producing visual patch tokens that are projected via a linear layer from dimension $768$ to $512$. In parallel, each physical transition tuple $(q_i, a_i^{\mathrm{cmd}}, q_{i+1}^{\mathrm{ach}})$ is concatenated and mapped through a shallow MLP, producing $512$-dimensional interaction tokens. We then concatenate the visual and interaction tokens and add learned temporal, spatial, and modality embeddings.

\begin{figure*}[t]
    \centering
    \includegraphics[width=\textwidth]{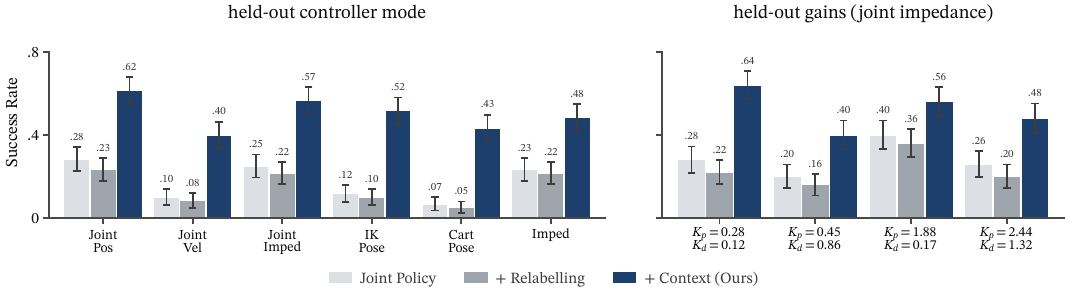}
    \caption{\textbf{Generalization across controller modes and gains.} Zero-shot success rates across six held-out controller modes ($N=60$, left) and four unseen $(K_p, K_d)$ joint impedance gains ($N=50$, right) compared to naive pooling and synthetic relabeling ($\pm\text{SE}$).}
    \label{fig:heldout_controllers}
\end{figure*}

\noindent\textbf{Interaction Bottleneck.} A fixed bank of $K = 256$ learned latent queries (dimension $512$) cross-attends over all visual and interaction tokens, distilling the variable-length multimodal sequence into 256 latent tokens. These tokens then pass through 6 latent self-attention blocks, allowing the latents to contextualize dynamic events across time. Finally, a linear layer projects the latents from dimension $512$ to $d_{\mathrm{VLA}}$, yielding compact interaction-context tokens:
\begin{equation}
    z_{\ell_s,c} = E_\phi(\mathcal{X}_{\ell_s,c}) \in \mathbb{R}^{256 \times d_{\mathrm{VLA}}}.
\end{equation}
This architecture addresses both challenges: cross-attention selectively pools sparse dynamic events, while the fixed $256$-token bottleneck encourages selective aggregation and keeps policy-side computation independent of context length.

\noindent\textbf{Pre-Computed KV-Caching for Real-Time Execution.} Because $\mathcal{X}_{\ell_s,c}$ is available before rollout, we run $E_\phi(\mathcal{X}_{\ell_s,c})$ once at episode start ($t=0$) and prepend $z_{\ell_s,c}$ to the VLA backbone ($\pi_{0.5}$~\cite{intelligence2025pi05}). Caching the key-value activations of $z_{\ell_s,c}$ only requires constant latency during closed-loop execution $\pi_\theta(a_{t:t+H} \mid o_t, \ell, \eta_c, z_{\ell_s,c})$, preserving real-time control regardless of context size.

\subsection{Data Augmentation}
\label{subsec:data_augmentation}

While dynamics remapping cannot synthesize new expert strategies at test time, it can still provide useful offline augmentation. Applying $R_{c \to c'}$ populates unobserved controller combinations with dynamically consistent actions, but changes only the commands used to realize a recorded motion—not the trajectory, contact pattern, or timing chosen by the demonstrator:
\begin{equation}
    R_{c \to c'}(\pi^\star_{\ell, c}) \neq \pi^\star_{\ell, c'},
    \label{eq:fundamental_inequality}
\end{equation}
where $\pi^\star_{\ell, c}$ denotes the expert demonstration policy under controller $c$. Consequently, trajectories relabeled to the same target controller can retain distinct modes of behavior. In wiping, for example, the same controller label may accompany both subsurface pressing inherited from compliant demonstrations and surface tracing inherited from stiff demonstrations. Behavioral context disambiguates which mode the policy should follow, making this multimodal augmentation usable during training.

\subsection{Policy Training}
\label{subsec:policy_training}
We train the policy backbone $\pi_\theta$ and context encoder $E_\phi$ jointly using conditional flow matching on action chunks. For each training step on task $\ell$ under controller $c$, we sample a target demonstration trajectory $\tau_{\ell,c} \in \mathcal{D}_{\ell,c}$ and a disjoint set of context trajectories $\mathcal{X}_{\ell,c} \subset \mathcal{D}_{\ell,c}$ under the same task and controller. We never encode the target trajectory itself. Note that encoder parameter sharing, trajectory disjointness, task diversity, and the fixed-size bottleneck encourage behavioral features that recur across tasks to emerge without requiring cross-task sampling. Cross-task context reuse is therefore not explicitly supervised; we evaluate it as an emergent generalization capability in Section~\ref{sec:experiments}.

Let $x_1 = a_{t:t+H}$ denote the demonstrated action chunk, let $x_0 \sim \mathcal{N}(0, I)$ denote standard Gaussian noise, and let $s \sim \mathcal{U}[0, 1]$ denote flow time. The linear probability path between noise and target actions is $x_s = (1 - s)x_0 + s x_1$ with target velocity field $u_s(x_s \mid x_1) = x_1 - x_0$. The final conditional flow-matching objective is:
\begin{equation}
\begin{aligned}
    \mathcal{L}(\theta, \phi) = \mathbb{E}_{s, x_0, \tau_{\ell, c}, \mathcal{X}_{\ell,c}} \Big[ \big\| &v_\theta\big(x_s, s \mid o_t, \ell, \eta_c, E_\phi(\mathcal{X}_{\ell,c})\big) \\
    &- (x_1 - x_0) \big\|^2 \Big].
\end{aligned}
\label{eq:flow_matching_loss}
\end{equation}

\begin{figure*}[t]
    \centering
    \includegraphics[width=\textwidth]{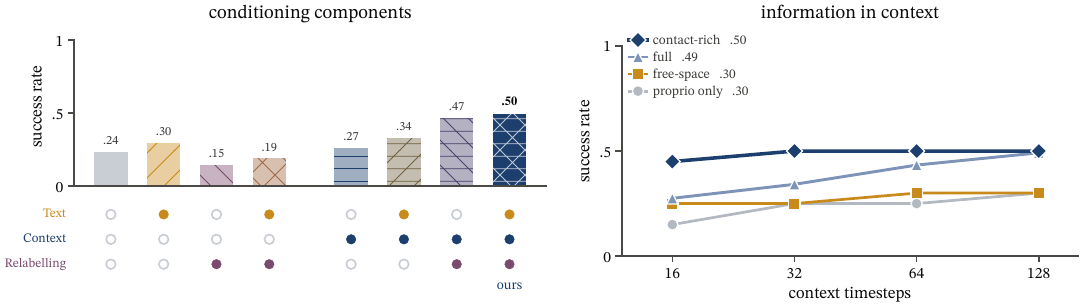}
    \caption{\textbf{Conditioning and information ablations.} Left: Behavioral context resolves the strategy multimodality introduced by relabeling ($0.15 \to 0.47$). Right: Focusing context on contact-rich transitions ($0.50$) captures the densest strategy signal, matching or exceeding uniform full-trajectory context ($0.49$).}
    \label{fig:information_ablations}
\end{figure*}

\section{Experiments}
\label{sec:experiments}

Our goal is to evaluate \emph{cross-controller transfer}: can in-context experience enable a single policy to adapt its execution strategy across diverse controllers and continuous gains? We seek the answer the following questions:

\begin{enumerate}
    \item \textbf{Is knowledge of controller dynamics sufficient for cross-controller behavioral adaptation, or does controller-conditioned expert behavior provide additional transferable information?}
    \item \textbf{Does adaptation generalize to unseen controllers and continuous gains?}
    \item \textbf{What information in context enables adaptation?}
    \item \textbf{Can interaction context transfer across tasks?}
\end{enumerate}

\noindent\textbf{Robots, Tasks, and Dataset.}
We evaluate on a WidowX manipulator setup ($50$\,Hz, wrist and third-person RGB cameras plus proprioception). Our evaluation spans five tasks: \emph{whiteboard wiping}, \emph{USB insertion}, \emph{faucet opening}, \emph{block throwing}, and held-out \emph{tool-bit insertion}. Demonstrations cover 4 source tasks $\times$ 6 controller types (joint position, joint velocity, IK pose, Cartesian pose, joint impedance, Cartesian impedance) $\times$ 4 gain settings per controller with 30 demonstrations per cell ($2{,}880$ trajectories total).

All methods share the same $\pi_{0.5}$ initialization, observations, demonstrations, and action horizon $H=20$; only conditioning and augmentation vary. We freeze DINOv2 and jointly train the context encoder and policy with Eq.~\eqref{eq:flow_matching_loss}. Behavioral context comprises $N=3$ deployment-controller trajectories, disjoint from the target trajectory, with $T=128$ sampled commanded--achieved transitions compressed into 256 tokens.

\noindent\textbf{Evaluation Protocol.}
We report binary success from matched initial-state and object-pose distributions. Success requires clearing the wiping target, fully seating the USB connector or tool bit, opening the faucet to its target configuration, or landing the block in the target region. Zero-shot excludes target-task demonstrations under the deployment controller; few-shot adds five. We run 20 trials per task--controller cell (60 per held-out mode when pooling three tasks), 50 per unseen gain pair, and 20 per context-reuse condition.

\noindent\textbf{Baselines.}
We compare methods that expose progressively richer information about the deployment controller. (1) \emph{Naive Joint Policy} pools demonstrations across controllers without controller identity. (2) \emph{Ground-Truth Controller Conditioning} conditions on the controller metadata $\eta_c$. (3) \emph{SysID} conditions on controller dynamics computed from calibration interactions. (4) \emph{Synthetic Relabeling} augments the dataset using torque-matched action remapping via Eq.~\eqref{eq:relabeling}. (5) \emph{Ours} conditions on both $\eta_c$ and behavioral context $z_{\ell_s,c}$ from expert trajectories on a source task under the deployment controller. (6) \emph{Same-Controller Upper Bound} uses target-task context collected under the deployment controller.

\section{Analysis}

\subsection{Is Controller-Dynamics Knowledge Sufficient for Behavioral Adaptation?}

We evaluate transfer to held-out task--controller combinations $\mathcal{D}_{\ell_t, c^\star}$ where the target task has no demonstrations under deployment controller $c^\star$ (Fig.~\ref{fig:cross_task_sysid}). This setting tests whether identifying controller dynamics alone is sufficient to recover expert performance. Ground-Truth Controller Conditioning and SysID both characterize how commanded actions produce physical motion, but neither reveals how an expert changes strategy under those dynamics. In contrast, our cross-task behavioral context—extracted from expert interaction on a different source task under $c^\star$—provides joint evidence of both physical controller response and controller-dependent demonstrator adaptation.

Across both zero-shot and few-shot regimes, behavioral context substantially outperforms dynamics identification alone. In zero-shot transfer, Ground-Truth Controller Conditioning ($12.0\%$) and SysID ($27.0\%$) struggle, while cross-task behavioral context reaches $50.0\%$ average success, approaching the same-controller upper bound ($75.0\%$). This margin persists with few-shot target data ($69.0\%$ vs.~$47.0\%$ for SysID and $34.0\%$ for Ground-Truth Conditioning). While empirical SysID improves over nominal parameters by calibrating motion response, predicting how a controller responds and predicting which actions an expert would choose under it remain fundamentally distinct problems. Without dense coverage across task--controller combinations, a policy cannot deduce the appropriate control strategy from dynamics alone. In contrast, cross-task behavioral context directly exposes demonstrator adaptation under the target dynamics providing priors on how to accomplish the task.

\begin{figure}[t]
    \centering
    \includegraphics[width=\columnwidth]{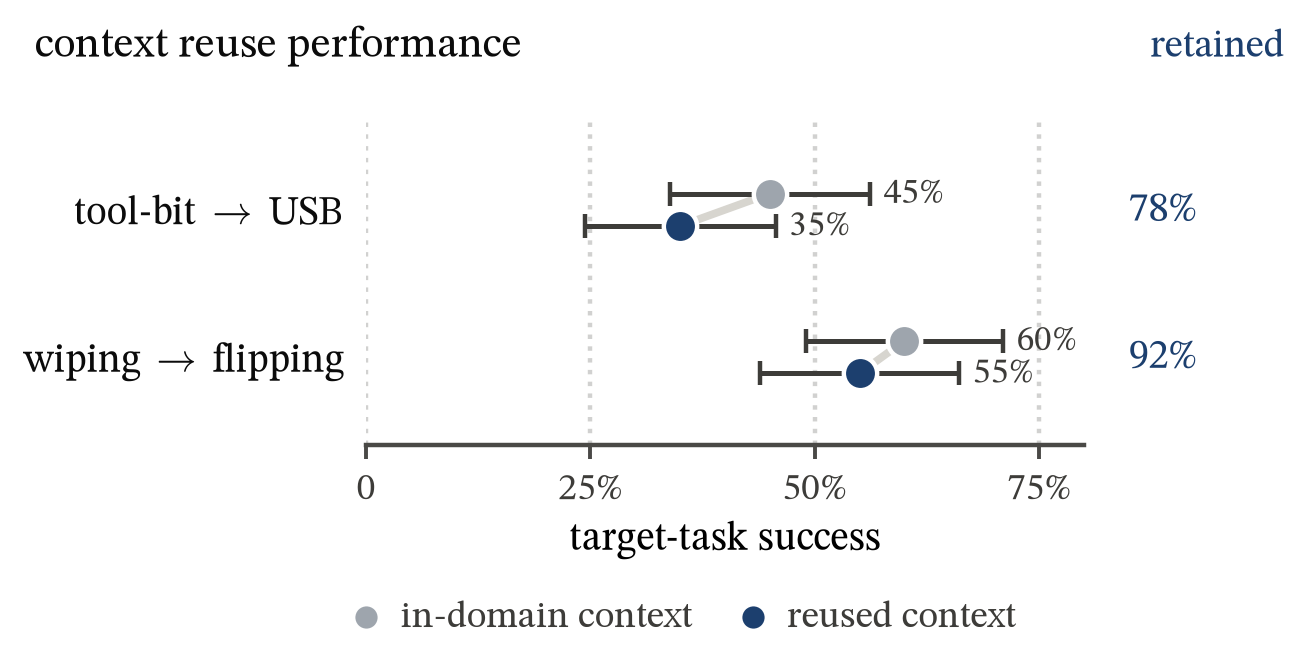}
    \caption{\textbf{Cross-task behavioral context transfer.} Reusing interaction-context tokens $z_{\mathrm{tool},c}$ extracted from tool-bit insertion to condition USB insertion under the same compliant controller achieves $35\%$ performance.}
    \label{fig:context_reuse}
\end{figure}

\subsection{Does Adaptation Generalize to Unseen Controllers?}

We next hold out the deployment controller entirely across all tasks (Fig.~\ref{fig:heldout_controllers}). In the discrete mode holdout (left), the naive joint policy achieves $17.5\%$ average success, while synthetic relabeling alone drops to $15.0\%$ due to multimodal action conflict under identical labels. Our context-conditioned policy reaches $50.3\%$ ($+32.8$ percentage points over the joint policy), nearly tripling performance. Context aids both kinematically exact modes (joint position $0.62$ vs.~$0.28$; velocity $0.40$ vs.~$0.10$) and contact-sensitive compliance modes (joint impedance $0.57$ vs.~$0.25$; Cartesian pose $0.43$ vs.~$0.07$; impedance $0.48$ vs.~$0.23$).

For continuous gain generalization under joint impedance control (right), we test four unseen $(K_p, K_d)$ pairs from $(0.28, 0.12)$ to $(2.44, 1.32)$\,kN/m. Context-conditioned adaptation achieves $52.0\%$ average success, significantly outperforming naive pooling ($28.5\%$) and relabeling ($23.5\%$). Performance peaks at $0.64$ under compliant gains (vs.~$0.28$ joint) and $0.56$ under stiff gains ($0.40$ joint). Rather than memorizing training gains, context supports continuous interpolation across the physical parameter plane.

\subsection{What Information in Context Enables Adaptation?}

Figure~\ref{fig:information_ablations} (left) evaluates conditioning over text, context, and relabeling, directly confirming the core insight of Section~\ref{subsec:dynamics_gap}: \emph{relabeling provides coverage, while behavioral context resolves multimodality}. Relabeling alone causes performance to decrease from $0.24$ to $0.15$ ($0.19$ with text), as torque remapping produces conflicting actions under identical controller labels. Behavioral context resolves this conflict, surging success to $0.47$ ($0.50$ with all conditioning active). Adding text yields only a $+0.03$ marginal gain, indicating that empirical interaction already exposes the behavioral regime.

Figure~\ref{fig:information_ablations} (right) investigates what information in context drives adaptation. An oracle isolating contact-rich transitions reaches $0.50$ success with minimal context. While uniform sampling across full trajectories begins lower due to the sparsity of contact events in unconstrained motion, scaling the context horizon enables it to match contact-rich performance ($0.49$). This confirms that the interaction bottleneck can extract behaviorally useful cues from uncurated demonstrations without manual phase segmentation, provided the horizon is long enough to observe contact interactions. In contrast, free-space motion lacks contact deflections and remains uninformative ($0.30$) regardless of context length, while proprioception alone cannot disambiguate environmental compliance from commanded intent ($0.30$).

\subsection{Can Interaction Context Transfer Across Tasks?}

Finally, we evaluate whether behavioral context extracted from one task can condition an entirely different task zero-shot under the same physical controller (Fig.~\ref{fig:context_reuse}). We extract context $\mathcal{X}_{\mathrm{tool},c}$ from demonstrations of contact-rich tool-bit insertion and compress it into latent tokens $z_{\mathrm{tool},c} = E_\phi(\mathcal{X}_{\mathrm{tool},c})$ to condition USB insertion under the same compliant controller. The reused context achieves $35\%$ success on USB insertion, retaining $78\%$ of the performance achieved with in-domain USB context. This result provides evidence that context containing source-task behavior can transfer useful controller-dependent cues. Qualitatively, the policy adopts the compliant search strategy learned from tool-bit insertion: upon encountering initial resistance at the port, the end-effector compliantly yields and maintains gentle surface contact while jiggling. Despite shifts in visual appearance and object geometry, the shared compliance strategy transfers across the two insertion tasks.

\section{Conclusion}
\label{sec:conclusion}
Our investigation demonstrates that the standard recipe for scaling generalist robot policies, pooling demonstrations across laboratories, breaks down at scale without considering controllers. Classical dynamics remapping and system identification fail to bridge this divide: matching instantaneous forces along an existing path can replicate motions, but cannot synthesize the compliant or exploratory strategies that are more robust under different dynamics. Our method compresses interaction histories through a fixed-size bottleneck, yielding a compact representation that can transfer useful behavioral evidence across distinct manipulation tasks. A key limitation of this approach is that analytical relabeling assumes impedance-style causality and may not directly extend to admittance controllers; furthermore, our conclusions reflect a single VLA architecture and laboratory setting.

\section {Acknowledgements}
This work was supported by the National Science Foundation under Award Nos. 2125511 and 2006388, by a Presidential Early Career Award for Scientists and Engineers (PECASE), and by the Office of Naval Research under Grant No. N00014-22-1-2621.

\bibliographystyle{IEEEtran}
\bibliography{references}

\end{document}